# CHARACTERIZING HUMAN BEHAVIOURS USING STATISTICAL MOTION DESCRIPTOR

Eissa Jaber Alreshidi[1] and Mohammad Bilal[2]

[1]University of Hail, Saudi Arabia; Comsats University, Pakistan

*ABSTRACT*

*Identifying human behaviors is a challenging research problem due to the complexity and variation of appearances and postures, the variation of camera settings, and view angles. In this paper, we try to address the problem of human behavior identification by introducing a novel motion descriptor based on statistical features. The method first divide the video into N number of temporal segments. Then for each segment, we compute dense optical flow, which provides instantaneous velocity information for all the pixels. We then compute Histogram of Optical Flow (HOOF) weighted by the norm and quantized into 32 bins. We then compute statistical features from the obtained HOOF forming a descriptor vector of 192-dimensions. We then train a non-linear multi-class SVM that classify different human behaviors with the accuracy of 72.1%. We evaluate our method by using publicly available human action data set. Experimental results shows that our proposed method out performs state of the art methods.*

## 1. INTRODUCTION

Automatic recognition of human actions play the importance role and is the most dominating research topic in computer vision research [1], [2], [3]. It has a wide range of application in automated surveillance [7], [8], [9] [10], human computer interaction [4], [5], [6], and video indexing [11], [12], [13] and video retrieval [14], [15], [16]. Human perform action for a specific purpose. For example, a patient is doing an exercise by interacting with environment by using his/her hands, arms, legs, and other body parts. An action can be either observed with the bare eyes or measured by using camera. With the bared eye, we can easily understand and classify that action into a specific category. For example, a person is walking or running, we can easily discriminate walking behavior from the running behavior.

For the video surveillance and smart rehabilitation, it is important to observe and monitor human actions for a long period [17], [18]. It is humanly impossible to monitor these actions for long duration due to limited human capabilities [19], [20]. Therefore, there is an increased interest to automate this process by using surveillance camera installed in different location of scene.

One of the ultimate goals of artificial intelligence research is to design a virtual machine [21], [22], [23] that can accurately analyze and understand humans' actions, so to reduce to human labor. For example, a patient is undergoing a rehabilitation exercise at home, and a virtual analyst that can understand and recognize his /her behaviors analyzes all his activities. With the help of such virtual analyst, we can prevent the patient from injuries. Such virtual analyst would be greatly beneficial as it saves the trips cost and medical cost. Other important applications including visual surveillance, entertainment, and video retrieval also need to analyze human actions in videos. Action recognition and prediction algorithms have wide range of applications.

Several work reported in literature [24], [25], [26], [27] with aim to automatically classify human actions that have substantially reduce the human labor in analyzing a large-scale of video data and provide understanding on the current state and future state.

Public safety and security is now becoming more important and in place under surveillance [28], certain human actions are not allowed [29]. In order to ensure public safety, surveillance camera are generally mounted on several places around the area under surveillance. With this camera network, action recognition and prediction algorithms may help in capturing malicious activities of human and therefore can reduce the risk caused by criminal actions. Therefore, acknowledging the importance of automatic human behavior analysis, several algorithms are proposed to solve this problem. Cutler et al. [30] [31] detect and recognize the periodic motion in very-low-resolution images. They first compute self-similarity, which evolves in time, and from this analysis, they showed whether an action is periodic. The problem with this method is that only used appearance features while computing similarity of appearance based features cannot discriminate the variation of posture and appearance between objects. Therefore, some of the researches consider the motion gradient information to classify actions. Efros et al. [32] introduced motion descriptor based on the optical flow and motion similarity. The measured noisy optical flow computed among the consecutive frames is smoothed out in four separated channels. Then a spatio-temporal motion descriptors is computed which then can classify using nearest-neighbor. They applied method on low-resolution videos and retrieve the postures of similar actions from action database. In [33], the method uses mid-level motion features, then threshold is used to extract low-level motion features classifier. Evaluation is performed with the dataset [34].. The motion features are similar to motion descriptors in [35]. The features are extracted using a variant of AdaBoost which focuses on the local regions. Other methods extract noisy motion through optical flow by using histograms of orientations. Chaudhry et al. [36] proposed a histogram of oriented Optical flow (HOOF) and used Binet-Cauchy kernels to classify human actions. One of the advantage of HOOF method is that it can alleviate the effect of noise, scale and motion variation. We propose an approach which is similar to HOOF but with significant differences: [37] computes histogram of optical flow (HOOF) from the time series data while our approach compute statistical features from the HOOF [38], [39], [40], which is very different from our approach.

Our main contribution are as follows: Our approach utilize dense optical flow information to build a motion descriptor, which can be used for identifying human behaviors. After computing optical flow, we build a motion descriptor by computing statistics from histogram of optical flow weighted by the norm of the velocity. The resultant statistics are concatenated representing motion descriptor, which are then used as the input of a SVM binary.

This paper is organized as follows: Section 2 introduces the proposed descriptor, section 3 demonstrates the effectiveness of the method and the last section concludes with a discussion and possible future works.

## 2. PROPOSED METHODOLOGY

The proposed methodology starts by computing a dense optical flow between two consecutive frames using the local jet feature space approach [41]. The advantage of computing dense optical flow is it allows us to segment the region, which contain motion information. After extracting foreground information, motion orientation histogram is then calculated, using typically 32 directions. Every direction bin is weighted by the norm of the flow vector. Finally, we compute list of statistics from HOOF, which will be our final motion descriptor. Later on, we train non-linear classifier that will classify different human action in different categories.

## 2.1 Optical Flow Computation

The first step to extract motion information is the computation of dense optical flow between two consecutive frames. For computing optical flow we employ methods [42] where gray value consistency, gradient constancy and smoothness in multi-scale constraints are used to compute highly accurate optical flow. Consider a feature point i in the frame associated to time t of a segment: its flow vector $Z_{i,t} = (X_{i,t}, V_{i,t})$ includes the location of feature and its velocity is represented by Vx and Vy. Where Vx represents the change in horizontal direction and Vy represents change in vertical direction. After computing optical flow for each pixel, we have now motion field where high magnitude represents the pixels corresponds to foreground and lower magnitude pixels represents the pixels corresponds to the background.

## 2.2 Particle Advection

After computing optical flow, the next step is to generate dense and long trajectories based on the optical flow [43], [44]. For doing so, we overlay grid of particles over the first optical flow field where each initial location (horizontal and vertical location) of the particle represents the source point. In order to generate dense trajectories, we keep the size of the grid as same as resolution of the frame. The size of particle is same as size of the pixel. This arrangement will incur computational costs. In order to reduce the computational cost and to generate dense trajectories, we reduce the resolution of the grid by dividing the size of the grid by a positive constant. During the advection process, we keep two separate flow maps, one to keep the horizontal coordinates and other map keeps track of vertical components of the trajectory. These map in general store the initial and subsequent positions of the point trajectories evolved during the process of particle advection.

The trajectories obtained through this process are suitable for structured crowds but in the unstructured crowds [45], [46], where people move in different directions, such trajectories do not represent the actual motion flow. The reason is that in unstructured crowds, the people move in arbitrary directions and in most of the case there is chance that particle will lose its path and become the part of different motion pattern moving towards a complete different direction. In this case, the trajectory is unreliable and erroneous. In order to avoid the above problem, we modify the above equation in the following way:

$$X_{(i,t+1)} = X_{(i,t)} + F(X_{(i,t)}) * B_i$$

Trajectories obtained using the above equation will avoid errors caused by particle drifting from one pedestrian flow to different motion pattern. The trajectories obtained through this method are precise, accurate but longer in length.

After particle advection, trajectories obtained corresponds to foreground while some of these trajectories correspond to the background of the scene or noise which are actually are not the part of actual motion pattern [47], [48], [49]. Therefore, in order to refine the obtained set of trajectories, we compute length of each trajectory by calculating the Euclidean distance between the source and sink points of the trajectories. We observed from our experiments that trajectories corresponds to noise and background are generally shorter in length. We exploit this information by setting a threshold value on the length of trajectories and suppress those trajectories whose length are shorter than the specified threshold

## 2.3 Computing Velocity Orientation Histogram

The next step is to obtain the distribution of orientation of each trajectory, since trajectory captures the spatial-temporal information; therefore, we need to estimate the distribution of orientations that can provide an aid in identifying type of motion. Let for non-zero vector $V$ computed between any two consecutive points of the trajectory. Let $\phi(V)$ denotes the quantized orientation. Similar to HOG descriptor [50], we compute the histogram of optical flow vectors of each trajectory weighted by the vector norm:

$$H_t(\omega) = \frac{\sum_{\{x:\ \phi(V_t(x))=\omega\}} \|V_t(x)\|}{\sum_{\{x:\ \|V_t(x)\|>0\}} \|V_t(x)\|}$$

Where $\omega \epsilon \{\omega o \dots \omega N-1\}$. Where $\omega N$ represents the number of orientations, which is set to 32 in our experiments. We capture the motion information similar to the HOOF descriptor of [51], except that the HOOF descriptor is not symmetrical. In other words, HOOF proposed [52] could not differentiate between the left and right directions while our proposed descriptor incorporate this information. Our proposed descriptor differentiate multiple directions and invariance to global motion information is addressed at the classification level.

## 2.4 Motion Descriptor

After computing velocity orientation histogram, we then compute statistical features that will capture important motion information. Let n represents the number of frames; we then compute statistics from the time series histogram of velocity orientations. Let $H t\ (\omega)$ represents the temporal histogram. We compute following features by using
$Ht\ (\omega)$:

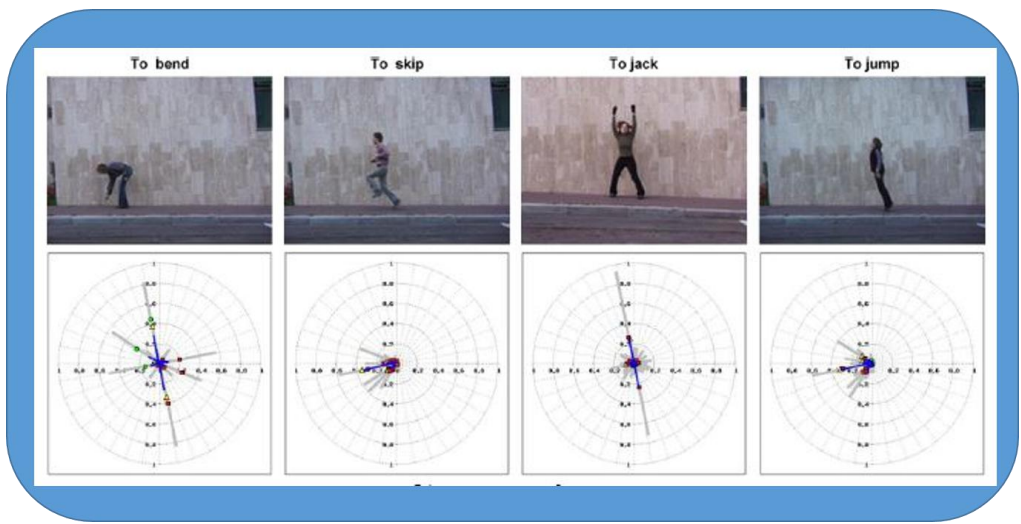

**Figure 1**: Examples of motion descriptor for different human behaviors

1. We compute the maximum value from the time series of histogram bins $H_t(\omega)$ in the following way

$$\max_{\{0 \leq t < n\}} \{H_t(\omega)\}$$

2. Mean of $H_t(\omega)$ is computed as

$$\sum_{\{0 \leq t < n\}} \frac{H_t(\omega)}{n}$$

3. Standard deviation of $H_t(\omega)$ is computed as

$$\sigma(\omega) = \sqrt{\sum_{\{0 \leq t < n\}} \frac{H_t^2(\omega)}{n} - \mu(\omega)^2}$$

We also divide the time interval into three segments, and compute the corresponding mean as follows. Let *n* is the total time interval

4. Mean for the first interval $\Psi_a(\omega)$ is formulated as

$$\sum_{\{0 \leq t < \frac{n}{3}\}} \frac{H_t(\omega)}{n/3}$$

5. Mean for the second interval $\Psi_b(\omega)$ is given by

$$\sum_{\{\frac{n}{3} \leq t < \frac{2n}{3}\}} \frac{H_t(\omega)}{n/3}$$

6. Mean for the third interval $\Psi_c(\omega)$ is given by

$$\sum_{\{\frac{2n}{3} \leq t < n\}} \frac{H_t(\omega)}{n/3}$$

Some examples of human behaviors and their corresponding motion descriptor are shown in Figure 1. Before computing motion descriptor, we divide the video into three temporal segments, i.e., and beginning. Middle and end segments [53]. We then compute motion descriptor for each segment of the video and concatenate all the three motion descriptors to represents over all motion behavior. For each motion descriptor, the blue lines represents the maximum values while gray lines represent the mean values. The red square represents mean values for the beginning of the video sequence; yellow triangle represents means values for the middle of the sequence while green disk represents the mean values associated to the end of video sequence [54]. From the Figure 1, it is obvious that our proposed motion descriptor is very discriminative and has the ability to identify distinct human behaviors.

## 2.5 Non-Linear Svm Classification

The SVM classifier applied in many pattern recognition problems. For classification, we use a non-linear support vector machine with a multi-channel kernel that efficiently combines multiple channels. We then define the multi-channel Gaussian kernel by:

$$K(H_i, H_j) = \exp\left(-\sum_{c \in C} \frac{1}{A_c} D_c(H_i, H_j)\right)$$

Where Hi = { hin } and Hj = { hjn } are the histograms for channel c and Dc (Hi, Hj) is the L2 distance defined as

$$D_c(H_i, H_j) = \frac{1}{2} \sum_{n=1}^{V} \frac{(h_{in} - h_{jn})^2}{h_{in} + h_{jn}}$$

Where V is the size of vocabulary. The parameter Ac is the average distances between all training samples for a channel c. For a given training set, we find the best set of channels C based on a greedy approach. We start with empty set of channels and add all possible channels. We then use greedy approach to evaluate each channel and remove channels until maximum is reached. In the case of multi-class classification, we use an approach of one-against all.

## 3. EXPERIMENTAL RESULTS

For the performance evaluation, we use dataset in [54]. The video database, as shows in Figure 2, is public available sequences, which contains 93 sequences of 10 human actions. The actions are: bend, jump, jack, jump forward- on-two-legs, jump-in-place-on-two-legs, run, gallop sideways, skip, walk, wave-two-hands, and wave one-hand) performed by 9 different actors. All the video sequences have the resolution of 180x144 pixels and are four seconds long video with average of 50 fps. This data set also include extracted foreground, obtained by background subtraction. We compute optical flow for only the foreground objects excluding the background in order to reduce the computation time.

For training, we use 2/3 of the dataset and the rest for testing. During our experiment, we randomly select six sequences in each action as a training set and the rest for testing.

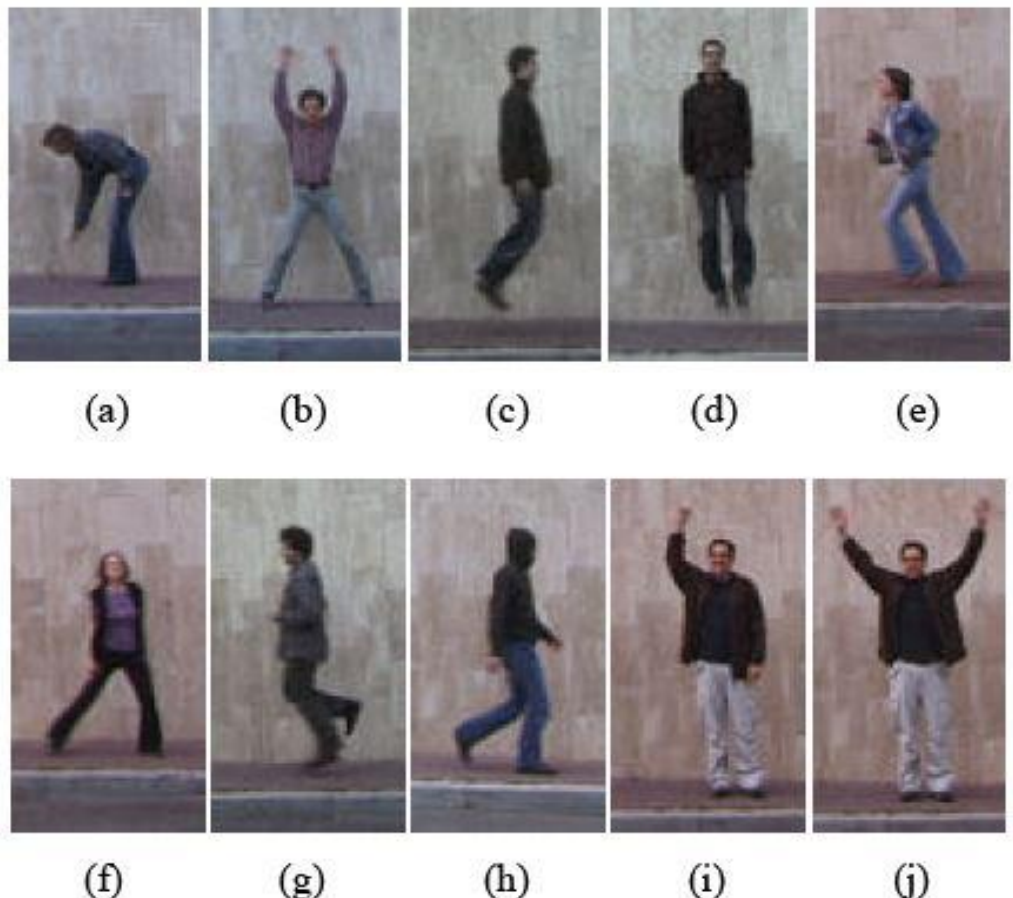

**Figure 2.** Sample frames from the dataset contain that 10 actions of 9 persons consists of (a) bend, (b) jack, (c) jump, (d) jump, (e) run, f) side, (g) skip, (h) walk, (i) wave1, (j) wave2

Figure 3, shows that confusion matrix for classification, where we train the classifier on one behavior and test it on the other behaviors. The average classification rate is 79.17%. There are some misclassified of bend with pjump because the person stand still before and after bending down which motion vectors is similar to the wrong category. In skipping, 3/4 sequences are classified as running. It is not unexpected because their motion and posture are very likely to each other. In addition, the misclassification of pjump with jump and wave2 with wave1 cause by similar pose too.

|       | Bend | Jack | Jump | Pjump | Run  | Side | Skip | Walk | Wave1 | Wave2 |
|-------|------|------|------|-------|------|------|------|------|-------|-------|
| Bend  | 0.87 | 0.00 | 0.00 | 0.33  | 0.00 | 0.00 | 0.00 | 0.00 | 0.00  | 0.00  |
| Jack  | 0.00 | 1.00 | 0.00 | 0.00  | 0.00 | 0.00 | 0.00 | 0.00 | 0.00  | 0.00  |
| Jump  | 0.00 | 0.00 | 1.00 | 0.00  | 0.00 | 0.00 | 0.00 | 0.00 | 0.00  | 0.00  |
| Pjump | 0.00 | 0.00 | 0.33 | 0.67  | 0.00 | 0.00 | 0.00 | 0.00 | 0.00  | 0.00  |
| Run   | 0.00 | 0.00 | 0.00 | 0.00  | 1.00 | 0.00 | 0.00 | 0.00 | 0.00  | 0.00  |
| Side  | 0.00 | 0.00 | 0.00 | 0.00  | 0.00 | 1.00 | 0.00 | 0.00 | 0.00  | 0.00  |
| Skip  | 0.00 | 0.00 | 0.00 | 0.00  | 0.75 | 0.00 | 0.74 | 0.00 | 0.00  | 0.00  |
| Walk  | 0.00 | 0.00 | 0.00 | 0.00  | 0.00 | 0.00 | 0.00 | 1.00 | 0.00  | 0.00  |
| Wave1 | 0.00 | 0.00 | 0.00 | 0.00  | 0.00 | 0.00 | 0.00 | 0.00 | 1.00  | 0.00  |
| Wave2 | 0.00 | 0.00 | 0.00 | 0.00  | 0.00 | 0.00 | 0.00 | 0.00 | 0.67  | 0.33  |

We also compare our method with other reference methods and the results are reported in Table 1. The other Classification methods perform leave-one out with nearest neighbor while we used hold out method. In our method, the sequences in training set are not used in the testing process. For Leave-one-out with X samples, the method is train all data except for one sample and test the prediction with the sample in each time X. The average error of X time is computed. So every data used to be a testing once and be a training X − 1 times. The variance of resulting evaluation also reduces as a number of training set increases. From the Table 1, it is obvious that our method outperforms other state-of-the-art methods.

| METHODS         | ERROR RATE |
|-----------------|------------|
| Khan et al [24] | **79.41**  |
| Ullah et al [43]| **95.24**  |
| Kong et al [3]  | **85.69**  |
| Proposed        | **72.11**  |

Signal & Image Processing: An International Journal (SIPIJ) Vol.10, No.1, February 2019

## 4. CONCLUSION

In this paper, we proposed an approach for recognizing human behaviors using our proposed features and non-linear SVM classifier. We demonstrated the capability of our approach in capturing the the dynamics of different classes by extracting these features. These features adopt the SVM to learn different classes. The main advantage of the proposed method is its simplicity and robustness.